\documentclass[lettersize,journal]{IEEEtran}
\usepackage{amsmath,amsfonts}
\usepackage{algorithm}
\usepackage{array}
\usepackage[caption=false,font=normalsize,labelfont=sf,textfont=sf]{subfig}
\usepackage{textcomp}
\usepackage{stfloats}
\usepackage{url}
\usepackage{verbatim}
\usepackage{graphicx}
\usepackage{orcidlink}
\usepackage{tikz}

\usepackage[numbers]{natbib}
\usepackage{bbm}
\usepackage{bm}
\usepackage{booktabs}
\usepackage{natbib}
\usepackage{multirow}
\usepackage{epstopdf}
\usepackage{hyperref}
\hypersetup{
    colorlinks=true,
    citecolor=blue,
    linkcolor=blue,
    urlcolor=blue
}
\usepackage{pifont}
\newcommand{\Checkmark}{\ding{51}}
\usepackage[utf8]{inputenc}
\usepackage{CJKutf8}

\usepackage{algorithm}  
\usepackage{algpseudocode}  
\usepackage{amsmath}

\begin{document}

\title{UMMAN: Unsupervised Multi-graph Merge Adversarial Network for Disease Prediction Based on Intestinal Flora}

\author{Dingkun Liu \and , Hongjie Zhou \and , Yilu Qu \and , Huimei Zhang \and , Yongdong Xu
\hspace{-1.5mm}$^{~\orcidlink{0000-0001-1234-1234}}$\textsuperscript{$\ast$}

\thanks{Dingkun Liu, IEEE Student Member, is with the University of Huazhong University of Science and Technology, 430074, China. Email: clarkjs0817@gmail.com.}
\thanks{Yongdong Xu is with the University of Harbin Institute of Technology, 264209, China. Email: ydxu@hit.edu.cn.}
\thanks{(Corresponding author: Yongdong Xu.)}}



\maketitle

\begin{abstract}
The abundance of intestinal flora is closely related to human diseases, but diseases are not caused by a single gut microbe. Instead, they result from the complex interplay of numerous microbial entities. This intricate and implicit connection among gut microbes poses a significant challenge for disease prediction using abundance information from OTU data. Recently, several methods have shown potential in predicting corresponding diseases. However, these methods fail to learn the inner association among gut microbes from different hosts, leading to unsatisfactory performance. In this paper, we present a novel architecture, \textbf{U}nsupervised \textbf{M}ulti-graph \textbf{M}erge \textbf{A}dversarial \textbf{N}etwork (UMMAN). UMMAN can obtain the embeddings of nodes in the Multi-Graph in an unsupervised scenario, so that it helps learn the multiplex association. Our method is the first to combine Graph Neural Network with the task of intestinal flora disease prediction. We employ complex relation-types to construct the Original-Graph and disrupt the relationships among nodes to generate corresponding Shuffled-Graph. We introduce the Node Feature Global Integration (NFGI) module to represent the global features of the graph. Furthermore, we design a joint loss comprising adversarial loss and hybrid attention loss to ensure that the real graph embedding aligns closely with the Original-Graph and diverges from the Shuffled-Graph. Comprehensive experiments on five classical OTU gut microbiome datasets demonstrate the effectiveness and stability of our method. (We will release our code soon.)
\end{abstract}


\section{Introduction}


\IEEEPARstart{A}{ccording} to incomplete statistics, there are at least 500 trillion microorganisms in the human body, outnumbering human cells by more than tenfold \citep{qin2010human}. Microorganisms are not only large in number, but also diverse in types, including viruses, bacteria, fungi, eukaryotes, etc. Collectively, these microbes constitute the human microbiome, an essential focus of contemporary biomedical research.

Some specific diseases are closely related to the abundance of microbial communities in the human body, with the vast majority of these microorganisms residing in the gut \citep{sender2016estimates}. As the largest microbial habitat in the human body, the gut microbiome has a direct impact on human health and diseases. For example, researchers have found significant differences in the proportions of Bacteroides and Actinobacteria in the intestines of obese and lean individuals, with the proportion of Firmicutes also influencing obesity \citep{dicksved2008molecular,ley2006human}. Various metabolites of gut microbes, such as short-chain fatty acid and aromatic amino acids, all directly or indirectly induce type 2 diabetes \citep{qin2012metagenome}. Patients with inflammatory bowel disease (IBD) exhibit a relatively disordered intestinal microbiota \citep{frank2007molecular,manichanh2006reduced}, with microbial metabolites such as acetate and butyrate playing crucial roles in immune regulation. The concentration of butyrate is notably decreased in patients with ulcerative colitis, highlighting the influence of gut microbes on IBD. In recent years, with the rise of 16SrRNA sequencing technology \citep{liu2020study}, new studies have shown that the number of lactic acid bacteria in the intestines of patients with irritable bowel syndrome decreased significantly, while the number of Veillonella in patients with constipation-predominant irritable bowel syndrome increased \citep{tana2010altered}. These findings further confirm the close connection between intestinal microbes and human health \citep{clemente2012impact,song2021review}.

With the rapid advancement of medical technology, obtaining relevant data has become increasingly convenient. 16SrRNA sequencing technology \citep{dethlefsen2008pervasive} can be used to obtain the abundance of human intestinal flora. Compared to whole genome sequencing, this technology is widely used due to its relatively low cost. However, processing this data to derive meaningful disease analysis results remains a significant challenge. Expert analysis of intestinal flora is complex and time-consuming, making it imperative to leverage artificial intelligence to enhance efficiency and reduce costs.

However, diagnosing diseases based on intestinal flora information has not achieved satisfactory results due to the complex and multiplex connections among the intestinal flora. Most existing approaches consider only a single relation-type even ignore the connection, fail to cover the intricate connection among the gut microbes of different hosts. Graph learning \citep{xia2021graph, tuan2021gnn, wang2024gnn} are well-suited for handling graph-structured data with rich relationships \citep{tena2022explainable, cheng2023gnndelete}, and can represent information at various depths. Therefore, we utilize Graph Neural Networks (GNNs) to learn the connections among gut microbes from different hosts, effectively guiding disease prediction. It is worth noting that our method does not rely on the labels in the process of obtaining the embeddings of the nodes and the graphs, which can obtain satisfactory embeddings in an unsupervised scenario.

Considering all the above challenges, we summarize the contributions of this paper as follows:

\begin{itemize}
   \item[$\bullet$] We are the first to introduce graph machine learning to the field of disease prediction based on gut microbiota. We propose a novel Unsupervised Multi-graph Merge Adversarial Network (UMMAN). It learns the intricate connection among intestinal flora of different hosts to guide the prediction of diseases.
   \item[$\bullet$] We use a variety of relation-types to build a Multi-graph, and shuffle the nodes of the corresponding graph to destroy the association among nodes. We introduce the adversarial loss and the hybrid attention loss as a joint loss in order to make the true embeddings agree with the Original-Graph and disagree with the Shuffled-Graph as much as possible.
   \item[$\bullet$] We propose the Node Feature Global Ensemble (NFGI) Descriptor, which includes node level stage and graph level stage to better represent the global embedding of a graph.
   \item[$\bullet$] Experiments on the benchmark datasets indicate that our UMMAN achieves state-of-the-art on the disease prediction task of gut microbiota, and also prove that our method is more stable than previous methods.
\end{itemize}

\section{RELATED WORK}
Benefiting from the rapid advancement of medical technology, many studies have been able to use OTU data for analysis \citep{edgar2013uparse,mcdowell2022machine}. Pasolli et al. \cite{pasolli2016machine} comprehensively evaluated the prediction tasks based on shotgun metagenomics and the method of microbial phenotype association evaluation, mainly using support vector machines and random forest models to predict diseases, and with the help of Lasso and elastic net (ENet) Regularized Multiple Logistic Regression. In their study, cirrhosis was the most predictive disease, and the model used in the study had good generalization ability for cross-stage data, but poor generalization ability for cross-datasets. Sharma et al. \cite{sharma2020taxonn} proposed TaxoNN to predict the conn between gut microbes and diseases, but only used two datasets to test the effect.

Manandhar et al. \cite{manandhar2021machine} used fecal 16S metagenomics data to analyze 729 IBD patients and 700 healthy individuals through 5 machine learning methods. After identifying 50 microorganisms with significant differences, the prediction was obtained through the random forest algorithm. Based on the data of the gut project in the United States, Linares et al. \cite{linares2021machine} used the glmnet model and the random forest model to predict the country of origin. Wong et al. \cite{wong2021analysis} investigated the possible gastrointestinal effects of neratinib in the treatment of breast cancer. By collecting stool samples from 11 drug-taking patients and classifying patients who may develop diarrhea by a tree-based classification method.

In general, at present, for data in the form of OTU datasets that are extended to table types, the traditional machine learning algorithm may be relatively more effective \citep{zhou2019review}. However, these algorithms are relatively fixed and there is not much space for improvement, reaching a bottleneck on the problem. More disappointing, basic CNNs perform well in many applications, but it can't beat the traditional machine learning algorithm in this case. This is because the basic CNN’s method cannot solve the problem of long-distance dependencies, nor can it adapt to data whose order can be changed at will. The exchange of rows and columns of the tabular datasets do not affect the results of pattern recognition, and more challengingly, diseases are not caused by a single gut microbe, but by a combination of a large number of microbial information. However, Graph Learning can help learn the intricate connection among the intestinal flora of different hosts. We propose the UMMAN unsupersived method, combining Graph Neural Network with gut microbiome for the first time, and it has achieved excellent performance on benchmark datasets.

\section{METHOD}
In this section, we will first introduce the overview of the architecture of our method --- UMMAN. In the following section, we introduce how to construct the Original-Graph and the Shuffled-Graph. We then present the details of the two stages of the Node Feature Global Integration descriptor: node-level stage and graph-level stage. Then we introduce the attention block to derive the embedding of each node. We also propose the joint loss which consists of adversarial loss $\mathcal{L}_{adv}$ and hybrid attention loss $\mathcal{L}_{h-attn}$ in the end.

\subsection{THE OVERVIEW OF UMMAN ARCHITECTURE}
The architecture of UMMAN we propose will be introduced detailedly in this part. As stated above, it exists close but extremely complex connection among gut microbes of different hosts, and such relation is implicit. Specifically, it is not the abundance of a single intestinal microbe that can establish a direct relationship with the final disease, but it is caused by the combined information of various microbes. For tabular datasets in the form of OTUs(A table with rows representing gut microbiota abundance and columns representing samples), their adjacent data are not correlated, so traditional convolutional neural networks are not applicable. Therefore, how to learn the connection among gut microbes of different hosts has become the biggest difficulty in this field, and we are the first to combine graph machine learning with gut flora disease prediction tasks which helps learn the connection. The following will introduce the architecture of UMMAN that we propose in detail. 

\begin{figure*}[t]
	\centerline{\includegraphics[width=0.9\linewidth]{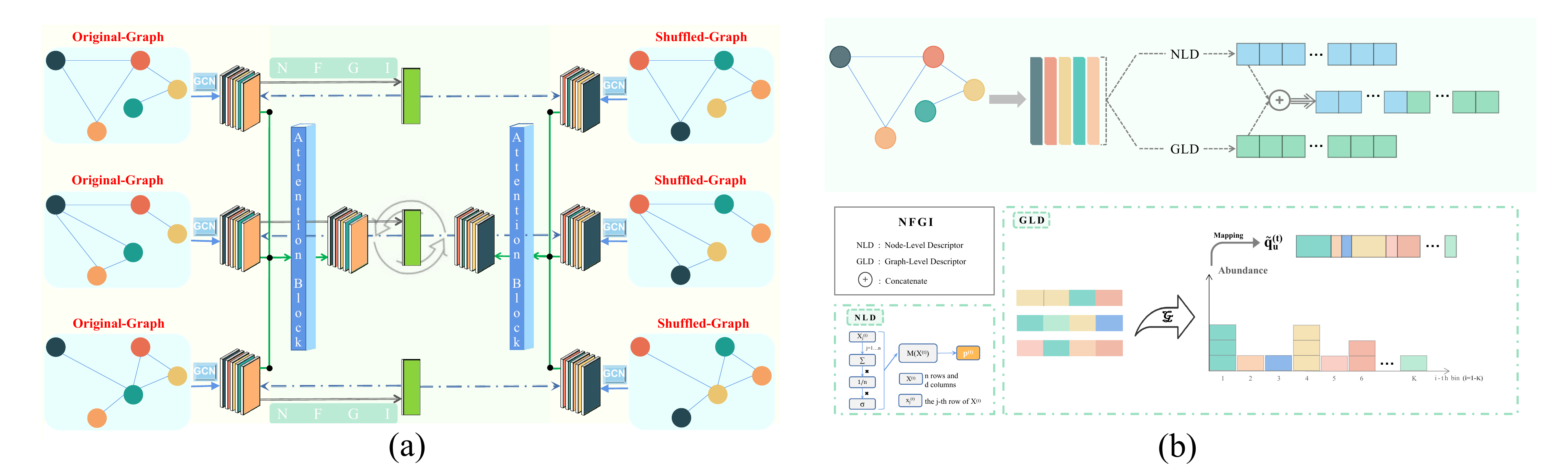}}
	\caption{The architecture of the UMMAN we propose. The subgraph on the left is the core architecture of the entire network. The subgraph on the right is the specific process of the NFGI module, including Node-level stage and Graph-level stage.}
	\label{figure1}
	\end{figure*}

The overview architecture of UMMAN is shown in \hyperref[figure1]{\textcolor{blue}{Fig .1}}. Each node represents a host. We use multiplex indicators to measure the similarity among nodes to build the Original-Graph. In order to make UMMAN learn associations more accurately, we destroy the Original-Graph's association among nodes to get Shuffled-Graph. The Original-Graph and Shuffled-Graph are updated through the Graph Convolutional Network at the same time in order to obtain the embeddings of each node. We introduce the attention block to obtain the embeddings of the nodes in the Original-Graph and the embeddings of the nodes in the Shuffled-Graph respectively. In addition, we design a Node Feature Global Integration (NFGI) descriptor to denote the embedding of a graph. Then, in order to make the true embedding that includes the complex and implicit relationships among gut microbes of different hosts agree with the Original-Graph as much as possible and disagree with the Shuffled-Graph as much as possible, we design the total joint loss function consists of an adversarial loss and hybrid attention loss, and used unsupervised methods to extract features from the Original-Graph. Finally, the Original-Graph after being processed by the attention block is fed into a dense layer classifier to produce the classification result.

\subsection{CONSTRUCT MULTI-GRAPH AND NODE EMBEDDING}
We find that the abundance data of a part of the intestinal flora in the dataset was extremely low in most hosts whose impact on the performance of the algorithm is detailed above. If all the data of the original dataset is retained, it may cause some intestinal flora to be almost completely absent in all hosts. We can consider such data as "dirty data", which will lead to training effects decline. Therefore, we removed the intestinal flora with low abundance in most hosts to improve the effect of feature extraction.

\begin{figure*}[t]
  \centerline{\includegraphics[width=0.9\linewidth]{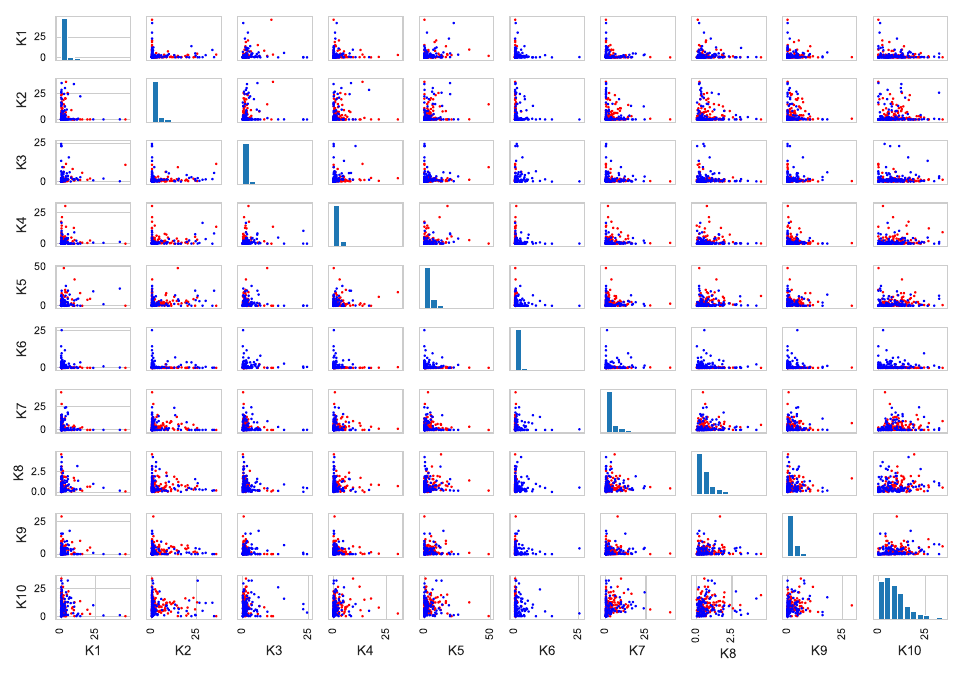}}
  \caption{Abundance correlations of the top 10 most abundant gut microbes in the Cirrhosis dataset.}
  \label{figure2}
  \end{figure*}

In order to make the motivation more convincing, we show the connection of the 10 most abundant flora in the cirrhosis dataset in the sample intestine in \hyperref[figure2]{\textcolor{blue}{Fig .2}}. The diagonal graph describes the distribution histogram of the flora, and the off-diagonal represents the relationships between the flora and the abundance in the intestinal tract of the sample, where the red dots represent healthy hosts, and the blue dots represent diseased samples. It can be found that the correlation is not obvious and appears to be disorganized, which proves that it is unreliable to measure the relation-type among hosts only based on the abundance of flora. To explore the implicit information of fused features in intestinal flora, we consider multiplex relation-types \citep{beals1984bray,jurman2009canberra} to measure the relationships among vectors during the initialization of building the edges among nodes in the graph:

\begin{equation}
  \mathcal{S}_{1}(m, n) = \frac{\sum\limits_{i=1}\limits^{d}|m_i - n_i|}{\sum\limits_{i=1}\limits^{d}m_i + \sum\limits_{i=1}\limits^{d}n_i} \\ 
\end{equation}

\begin{equation}
  \mathcal{S}_{2}(m, n) = \sqrt{\sum\limits_{i=1}\limits^{d} (m_i - n_i) ^ 2} \\ 
\end{equation}

\begin{equation}
  \mathcal{S}_{3}(m, n) = \sum\limits_{i=1}\limits^{d} \frac{|m_i - n_i|}{|m_i| + |n_i|} ,\\ 
\end{equation}

  where Bray Curtis Distance, Euclidean Distance and Canberra Distance are used to initialize a graph. We construct the original graph by measuring the multivariate discrepancy between nodes, two nodes are considered to be connected if the discrepancy among their embeddings is below a variable threshold $\theta$. In order to make our model more robust, and to obtain the correlation among nodes more reliably, we design the adversarial control group, i.e., keep the position of the edges unchanged, randomly disrupt the nodes to get the Shuffled-graphs, and train the discriminator to get the adversarial loss. The specific process will be introduced in detail later.


  We introduce an encoder module for each relation type inspired by the Graph Convolutional Network\citep{kipf2016semi}, aiming to obtain the embedding of each node in the graph. We define a conditional function $\mathcal{F}$ as an update function between layers, the Shuffled-Graph is operated by the same process $\mathbbm{R}^{N\times F} \hookrightarrow \mathbbm{R}^{N\times D}$:

\begin{align}
\mathcal{X}_{j}^{(l + 1)} &= \mathcal{F}(\mathcal{X}_j^{(l)}, \hat{A} | W^{(l)}, b^{(l)}) \nonumber \\
&= \sigma \left( \sum_{j \in N_i} \hat{\mathcal{D}}^{- \frac{1}{2}} \hat{\mathcal{A}} \hat{\mathcal{D}}^{- \frac{1}{2}} \mathcal{X}_j^{(l)}W^{(l)} + b^{(l)} \right),
\end{align}

  where $\mathcal{X}_j^{(l)}$, is the embedding of the l layer of the node whose index is j, $N_i$ represents the set of nodes adjacent to $\mathcal{X}_j^{(l)}$, $\hat{\mathcal{A}}$ and $\hat{\mathcal{D}}$ are the adjacency matrix and degree matrix of a certain graph, $W^{(l)}$ and $b^{(l)}$ are the trainable parameters of the l-th layer's weight matrix and bias, $\sigma$ is the nonlinearity layer which is designed as ReLU in our method. Through Graph Convolutional Network, ${X}_j^{(t)}$ become a D-dimensional tensor representing the embeddings of the node with index j of the t-th relation-type.

\subsection{NODE FEATURE GLOBAL INTEGRATION DESCRIPTOR WITH TWO STAGES}
In the NFGI module, we propose a novel contribution to characterize a graph $\mathcal{J}^{(t)}  / \mathcal{\tilde{J}}^{(t)}$ with two stages($\mathcal{J}^{(t)}$ is the Original Graph constructed according to the t-th relation-type, and $\mathcal{\tilde{J}}^{(t)}$, is the corresponding graph after being shuffled, since the operations of the two graphs are equivalent, we use $\mathcal{J}^{(t)}$ uniformly in the following text). In the conventional method, only the features of each node are fused, but using this method to characterize a graph only considers the information of each node. Therefore, our NFGI module adopt a two-stage process, i.e., node-level stage and graph-level stage to jointly characterize the features of the graph to better obtain the embedding.

$\textbf{\textit{Node-Level stage}}$. We first compute the node-level contribution to graph $\mathcal{J}^{(t)}$ based on the characteristics of each node. That is, to fuse the embedding of each node after the Graph Convolutional Network. We employ a function $\mathcal{M}$ : $\mathbbm{R}^{N\times D} \hookrightarrow \mathbbm{R}^D$:

\begin{equation}
  p^{(t)} = \mathcal{M} (X^{(t)}) = \sigma (\frac{1}{N}\sum\limits_{i=1}\limits^{N}X_{i}^{(t)}). \\ 
\end{equation}

$p^{(t)}$ summarizes the node-level features of the graph $\mathcal{J}^{(t)}$. Among them, $X^{(t)}$ is a matrix with n rows and d columns, $X_{i}^{(t)}$ is the i-th row of matrix $X^{(t)}$ and $\sigma$ denotes the logistic sigmoid nonlinearity function. 

$\textbf{\textit{Graph-Level stage}}$. In the graph-level stage, we concatenate the embedding of each node to build a tensor, and then flatten the tensor to get the total information of the whole graph which can form K bins. Let $\{{x_j^{*}}\}_{j = 1...n}$ be the each center of the histogram bins. The value included in each bin is within the range of r centered on $\{{x_j^{*}}\}$. Given a certain feature value $x_i$, we define a function $\Psi$, the index $\Psi(x_{i}^{(t)})$ corresponding to the bin that the value belongs to. On the whole, the function $\mathcal{G}$ denotes the embedding of the graph in the graph-level stage: $\mathbbm{R}^{N\times D} \hookrightarrow \mathbbm{R}^K$:

\begin{equation}
  \tilde{q}_{u}^{(t)}  = \mathcal{G} (X^{(t)}) = \sum\limits_{i=1}\limits^{N\times D}C(|x_{i}^{(t)}|)\delta[\Psi(x_{i}^{(t)})-u]\\ 
\end{equation}

\begin{equation}
  q^{(t)} = \mathcal{E} [\tilde{q}_{u = 1...K}^{(t)} ],\\ 
\end{equation}

where $\delta$ is the Kronecker delta function. The normalization constant C is derived by imposing $\sum\limits_{i=1}\limits^{N \times{D}} \tilde{x}_{i}^{(t)} = 1$, performing data smoothing on the graph-level results. The graph-level embedding result can describe the global feature of the graph. Therefore, after concatenating each graph-level stage feature using the symbol $\mathcal{E}$ , the final graph-level embedding result is obtained: $\mathbbm{R}^{N\times D} \hookrightarrow \mathbbm{R}^{D + K}$. Finally, we concatenate the node-level embedding and graph-level embedding by the axis 0.  

\begin{equation}
  G^{(t)} = \mathcal{E} [\tilde{p}^{(t)} , \tilde{q}^{(t)}] \\ 
\end{equation}

\subsection{CONVERGE MULTI-GRAPH BY ATTENTION}
Single-scale CNN fail to capture the intricate relationships among intestinal flora, while conventional multi-scale CNNs sometimes fall short in effectively integrating information from various scales using concatenation-based approaches.

The significance of self-attention CNNs in amalgamating multi-scale information is due to several factors. To begin with, the self-attention mechanism in CNNs improves classifying capability by selectively concentrating on significant regions or features in input data, consequently increasing classification accuracy and aiding in detecting nuanced patterns. Moreover, self-attention CNNs generate interpretable outcomes by indicating the regions that most influence the classification decision. Additionally, self-attention CNNs perform exceptionally well in managing high-dimensional data, allowing them to learn distinct features in multiple dimensions, thereby capturing intricate patterns and dependencies in gut microbiome abundance data. In general, self-attention is a notable subtype of attention mechanisms because of its ability to capture long-range dependencies, adaptively weigh input elements, manage variable-length sequences, ensure interpretability, and combine with other deep learning techniques. In this part, we utilize it to merge Original-graph and Shuffled-graphs that have been updated by Graph Convolutional Network.

After using Graph Convolutional Network to update the Original-graph and Shuffled-graphs, we get the embedding of each node. Each node in the Multi-graph $\mathcal{J}^{(t)}$ gets a feature vector to represent, and then we use  attention block to update the embeddings of the corresponding nodes \citep{vaswani2017attention} of the Multi-graph to an embedding result representing the node feature. The attention block is defined as follows:


\begin{align}
\mathbf{X}_{\text{attn}} &= \mathcal{A}\text{ttn}(X_i^{(t)} \mid t \in \mathcal{T}) \nonumber \\
&= \sum_{t \in T} \frac{\exp(\text{query}^{(t)}X_i^{(t)})}{\sum_{t' \in T'} \exp(\text{query}^{(t')}X_i^{(t')})} X_i^{(t)},
\end{align}


where $\text{query}^{(t)} \in \mathbbm{R}^{D + K}$ is the feature vector of relation-type $t$. The Original Graphs and the Shuffled Graphs can obtain the embedding results $\bm{x}_i$ and $\tilde{\bm{x}}_i$ of each node after passing through the attention block. Finally, the embedding of nodes can be sent to the MLP for prediction.

\subsection{LOSS FUNCTIONS}
The loss functions of our model consists of an adversarial loss $\mathcal{L}_{adv}$  and hybrid attention loss $\mathcal{L}_{h-attn}$.
The total loss function is as follows:

\begin{equation}
  \mathcal{L} = \mathcal{L}_{adv} + \grave{\eta} \mathcal{L}_{h-attn}, \\ 
\end{equation}

where $\grave{\eta}$ is a learnable coefficient for the loss terms.

\subsubsection{ADVERSARIAL LOSS}
We not only use a variety of relation-types to construct Multi-graph and obtain the embedding of each Node through Graph Convolutional Network and Attention Block, but also the Original-Grpah Randomly shuffle to break the correlation among nodes. Therefore, we calculate the positive correlation loss between the global embedding obtained by NFGI (Node Feature Global Integration Descriptor) and the embedding of the nodes of the Original-Graph obtained by each relation-type. At the same time, on the converse side, we calculate the negative correlation loss between the global embedding and the embedding of the nodes of the Shuffled-Graph, and define this joint loss as Adversarial Loss inspired by \cite{pajot2018unsupervised}. In other words, we calculate the adversarial loss from the embedding obtained from the constructed Original-Graph (Positive) and the Shuffled-Graph (Negative) that destroys the correlation among nodes, which is defined as:

\begin{align}
\mathcal{L}_{adv} = &\sum_{t \in \mathcal{T}} \sum_{i=1}^{N} \log \sigma \left((H^{(t)})^\mathrm{T} W^{(t)}X_i^{(t)}\right) \nonumber \\
&+ \sum_{t \in \mathcal{T}} \sum_{j=1}^{N} \log \left(1 - \sigma \left((H^{(t)})^\mathrm{T} \tilde{W}^{(t)}\tilde{X}_i^{(t)}\right)\right).
\end{align}

\subsubsection{HYBRID ATTENTION LOSS}
The Hybrid Attention Loss we proposed comprehensively considers the embedding after the nodes of the Original-Graph and the Shuffled-Graph pass through the attention block. The Hybrid Attention Loss $\mathcal{L}_{h-attn}$ can make the global embedding matrix of real graph agree with $\bm{X}_{attn}^{(t)}$, and disagree with $\bm{\tilde{X}}_{attn}^{(t)}$, thereby improving the confidence of the attention block. The Hybrid Attention Loss function is defined as:

\begin{equation}
  \mathcal{L}_{h-attn} = (\mathcal{P} - \bm{X}_{attn}) ^ 2 - (\mathcal{P} - \tilde{\bm{X}}_{attn}) ^ 2. \\ 
\end{equation}

\begin{algorithm}[H]
\caption{Unsupervised Multi-graph Merge Adversarial Network}
\label{alg:NNModel}
\begin{algorithmic}
\Require
feature $\mathbf{F}$, origin data $\mathbf{O}$,\
shuffled data $\mathbf{S}$, graph num $\mathbf{N_g}$, head num $\mathbf{N_h}$
\Ensure
loss $\mathcal{L}$
\For{$i = 1$ to $\mathbf{N_g}$}
\State $\mathbf{pos} \gets \mathbf{GCN}(\mathbf{F[i]}, \mathbf{O[i]}, i)$
\State $\mathbf{p} \gets \mathbf{NFGI}(\mathbf{pos})$
\State $\mathbf{neg} \gets \mathbf{GCN}(\mathbf{S[i]}, \mathbf{O[i]}, i)$
\State $\mathcal{L}_{adv} \gets \mathbf{Disc}(\mathbf{p}, \mathbf{pos}, \mathbf{neg})$
\State $\mathbf{P.append(pos)}$
\State $\mathbf{N.append(neg)}$
\EndFor
\For{$h = 1$ to $\mathbf{N_h}$}
\State $\mathbf{P_{attn}}, \mathbf{N_{attn}}\gets\
\mathbf{Attn[h]}(\mathbf{P}, \mathbf{N})$
\EndFor
\State $\bm{X}_{attn} \gets \mathbf{mean}(\mathbf{P_{attn}})$
\State $\tilde{\bm{X}}_{attn} \gets \mathbf{mean}(\mathbf{N_{attn}})$
\State $\mathcal{L}_{pos} \gets (({\mathcal{P}} - \bm{X}_{attn})^2).\mathbf{sum}()$
\State $\mathcal{L}_{neg} \gets (({\mathcal{P}} - \tilde{\bm{X}}_{attn})^2).\mathbf{sum}()$
\State $\mathcal{L}_{h-attn} \gets \mathcal{L}_{pos} - \mathcal{L}_{neg}$
\State $\mathcal{L} \gets \mathcal{L}_{adv} + \mathcal{L}_{h-attn}$
\State \textbf{return} $\mathcal{L}$
\end{algorithmic}
\end{algorithm}

\section{EXPERIMENT}
\subsection{SET UP}
We performe extensive experiments on five real-world datasets to evaluate the performance of our UMMAN model. We also compare it to several state-of-the-art
machine learning and deep learning models. In pre-processing, we removed the intestinal flora which abundance lower than 0.01 in most hosts (set to 120). During the initialization of the graph, set the threshold $\theta$ to 0.6. We randomly divide each dataset into five folds, with four folds for the training set and one fold for the validation set, using the k-fold method for cross-validation. We randomly select five different random seeds for each validation dataset and take the average performance as the final result. We use a three-layer Graph Convolutional Network to extract the embedding for each node. In training, we use the Adam optimiser with an initial learning rate of 0.001 and epochs of 1000. We set the node embedding dimension to 256. 
\subsection{DATASETS}
Five classical gut microbiota datasets are used for the experiments. We use five available disease-associated metage nomic datasets spanning four diseases: liver cirrhosis, inflammatory bowel diseases (IBD), obesity, and type 2 diabetes in Asia and Europe, respectively. These datasets have recorded the abundance of 1331 gut microbes in the sample intestine, as well as information such as gender, age, and region.

$\textbf{\textit{Cirrhosis}}$ \citep{qin2014alterations}. Qin et al. extracted total DNA libraries from fecal samples of patients with cirrhosis and healthy controls, which can characterize the gut microbiome of patients with cirrhosis. Specifically, they used Illumina HiSeq 2000 for sequencing, which produced an average of 4.74 Gb of high-quality sequences per sample, and a total of 860 Gb of 16SrRNA gene sequence data.

$\textbf{\textit{IBD}}$ \citep{qin2010human}.  Qin et al. collected stool samples from volunteers and performed Illumina GA sequencing. All Reads were assembled using Soapnovo19. Using BLAT36 to construct a non-redundant gene set for inflammatory bowel disease by pairwise comparison of all genes. Each sample yields an average of 4.5Gb of high-quality sequences.

$\textbf{\textit{Obesity}}$ \citep{le2013richness}. Obesity is one of the most serious (proportional) gut microbiota-related diseases facing the world. Le et al. extracted the abundance of more than a thousand species of flora in the gut by sequencing fecal samples from Obesity patients and thinner stature.

$\textbf{\textit{T2D}}$ \citep{qin2012metagenome}. Stool samples are collected from patients with type 2 diabetes and healthy controls in order to obtain the DNA data of the flora in the samples, using the whole genome sequencing method, and then sequenced all the DNA samples, and analyzed the intestinal microbial DNA of 345 Chinese Two-stage MGWAS was performed for deep shotgun sequencing, with an average of 2.61 Gb per sample and a total of 378.4 Gb of high-quality DNA data.

$\textbf{\textit{WT2D}}$ \citep{karlsson2013gut}. Shotgun sequencing is used to analyze the whole genome sequence of fecal samples from European women, sequenced on Illumina HiSeq 2000, obtained an average of 2Gb of sequencing data per sample, and a total of about 449 Gb of data. Different from T2D, the target of T2D dataset is Chinese, while WT2D is sampled in Europe.

\subsection{COMPARISON WITH EXISTING WORK}
We carry out quantitative comparisons on the results between our method and existing work on five datasets. The Random Forest and Support Vector Machine can be used in the disease prediction of intestinal flora \citep{pasolli2016machine}, although the results obtained are acceptable, the performance of such algorithms has reached the bottleneck so that it is difficult to improve. The basic CNN series methods are particularly effective at extracting closely located features, but they encounter difficulty in capturing attention over long distances. Additionally, due to the property of uncorrelated adjacent data in the OTU tabular dataset, traditional convolutional neural networks struggle to extract the intrinsic connections among gut microbiota. We innovatively combine a graph machine learning algorithm with the gut microbiota disease prediction task. We first compared the performance results of our method and the previous 8 methods on the Acc and AUC indicators on the benchmark datasets, as shown in \hyperref[table1]{\textcolor{blue}{Table 1}} (bold indicates the best performance, underline indicates the second-best performance).

\begin{table}[!htbp] 
	\caption{Comparison of our method with previous methods on the Accuracy and the AUC metrics on the benchmark datasets.} 
	\resizebox{\linewidth}{!}{
	\centering
  \fontsize{16}{23}\selectfont 

	\begin{tabular}{cccccccccccc}
		\toprule
		\multirow{2}*{Method} & \multicolumn{2}{c}{Cirrhosis} & \multicolumn{2}{c}{IBD} & \multicolumn{2}{c}{Obesity} & \multicolumn{2}{c}{T2D} & \multicolumn{2}{c}{WT2D}\\
		\cmidrule(lr){2-3}\cmidrule(lr){4-5}\cmidrule(lr){6-7}\cmidrule(lr){8-9}\cmidrule(lr){10-11}
		& Acc & AUC & Acc & AUC & Acc & AUC & Acc & AUC & Acc & AUC\\
		\midrule 
		RF \citep{pasolli2016machine}  & \underline{0.877} & \underline{0.945} & 0.809 & \underline{0.890} & 0.644 & \underline{0.655} & \underline{0.664} & \underline{0.744} & \underline{0.703} & \underline{0.762}\\
		SVM \citep{pasolli2016machine} & 0.834 & 0.922 & 0.809 & 0.862 & 0.636 & 0.648 & 0.613 & 0.663 & 0.596 & 0.664 \\
		FC-NN \citep{rumelhart1986learning} & 0.847 & 0.936 & 0.821 & 0.874 & \underline{0.645} & 0.647 & 0.656 & 0.705 & 0.657 & 0.71 \\
		FC-NN+SVM \citep{rumelhart1986learning, pasolli2016machine} & 0.836 & 0.917 & \underline{0.833} & 0.889 & 0.609 & 0.614 & 0.637 & 0.673 & 0.675 & 0.73 \\
		CNN \citep{krizhevsky2012imagenet} & 0.837 & 0.922 & 0.826 & \underline{0.890} & 0.629 & 0.635 & 0.638 & 0.688 & 0.676 & 0.738 \\
		CNN+RF \citep{krizhevsky2012imagenet, breiman2001random} & 0.844 & 0.907 & 0.819 & 0.832 & 0.62 & 0.601 & 0.637 & 0.681 & 0.693 & 0.725 \\
		TaxoNN \citep{sharma2020taxonn}  & 0.842 & 0.911 & 0.826 & 0.887 & 0.629 & 0.627 & 0.646 & 0.733 & 0.688 & 0.745 \\
		GraphSAGE \citep{hamilton2017inductive}  & 0.826 & 0.877 & 0.818 & 0.875 & \underline{0.645} & 0.647 & 0.656 & 0.705 & 0.657 & 0.71 \\
		UMMAN(Ours)  & \textbf{0.886} & \textbf{0.954} & \textbf{0.889} & \textbf{0.930} & \textbf{0.706} & \textbf{0.684} & \textbf{0.748} & \textbf{0.777} & \textbf{0.867} & \textbf{0.797} \\
		\bottomrule
	\end{tabular}
	}
     \label{table1}
\end{table}


\begin{figure}[t]
	\centering
	\centerline{\includegraphics[width=0.9\linewidth]{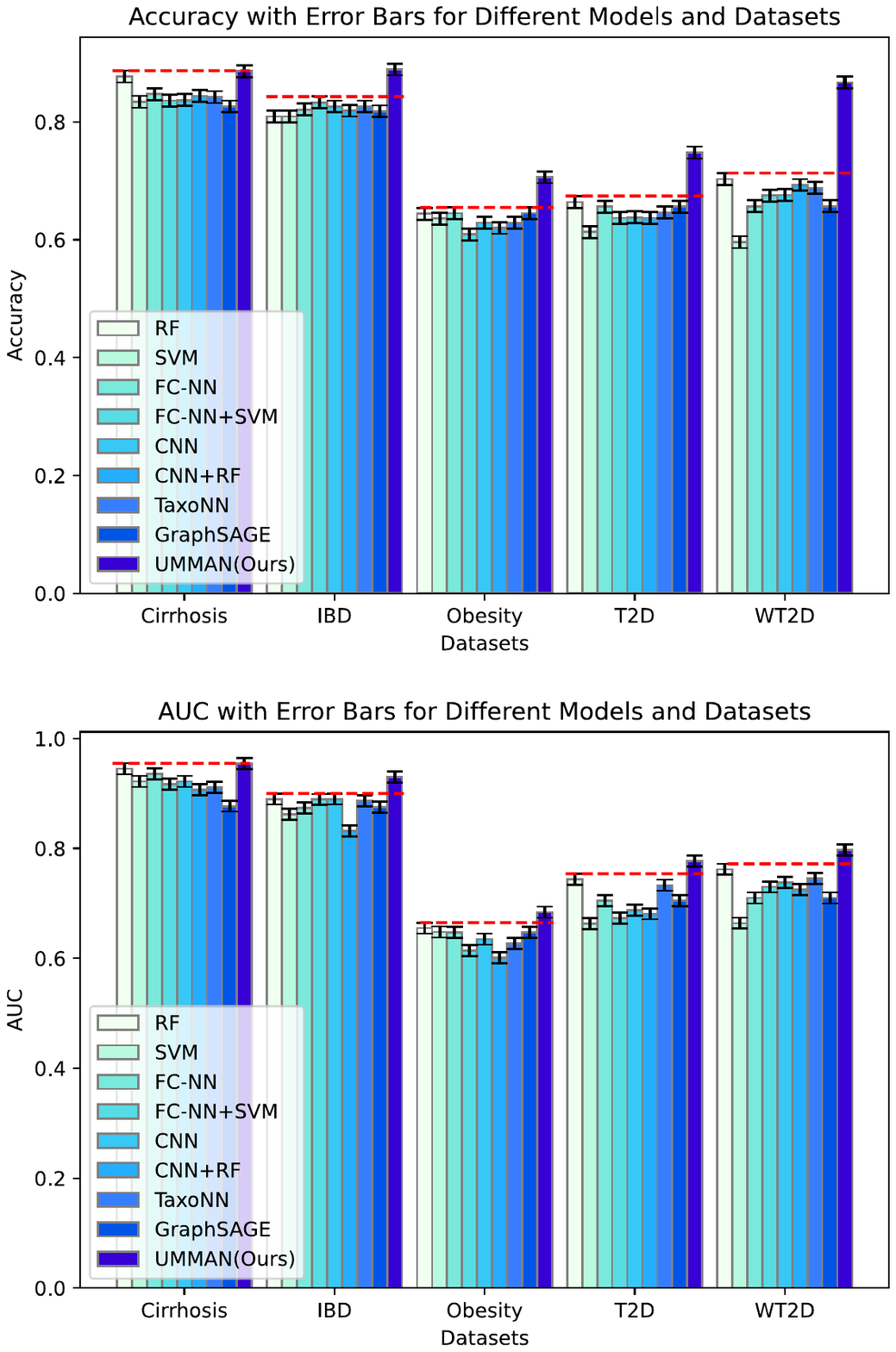}}
	\caption{Intuitive comparison of our method with previous work on the five OTU datasets.}
	\label{figure3}
	\end{figure}

\begin{figure*}[ht]  
    \centering
    \centerline{\includegraphics[width=1\linewidth]{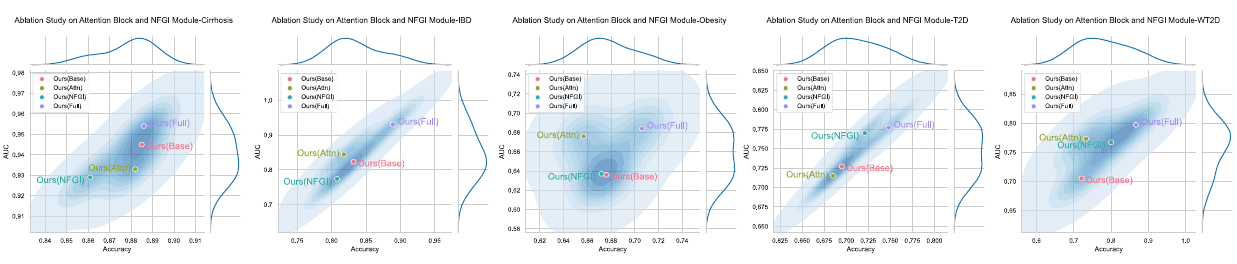}}
    \caption{The comparison on four variants of our method.}
    \label{figure4}
\end{figure*}

Our method, UMMAN, achieves the best performance on the metrics across the five datasets on all indicators. From the experimental results in \hyperref[table1]{\textcolor{blue}{Table 1}} and \hyperref[figure3]{\textcolor{blue}{Fig .3}}, it can be found that almost all the optimal and second-best results are derived from our method and Random Forest. This may be due to the robustness of traditional machine learning algorithms. Therefore, in order to prove that our method (UMMAN) not only achieves SOTA on the result value, but also surpasses the previous methods in all aspects in terms of stability and other indicators. \hyperref[table2]{\textcolor{blue}{Table 2}} compares the performance of our method and traditional machine learning algorithms on five datasets in detail. We use K-fold cross-validation on five datasets of \textbf{\textit{Cirrhosis}}, \textbf{\textit{IBD}}, \textbf{\textit{Obesity}}, \textbf{\textit{T2D}}, \textbf{\textit{WT2D}}, and use Accuracy, Precision, Recall, F1-Score, AUC indicators to compare with previous methods in an all-round way. \hyperref[figure4]{\textcolor{blue}{Fig .4}} shows the visual comparison results of the four variants of the ablation experiment on five gut microbiome datasets. Each line represents a contour line, and the horizontal and vertical coordinates represent Accuracy and AUC metrics, respectively. In short, the method closer to the upper right region has better performance. The ablation experiments show that our method achieves the best results in almost all results, and the stability performs well, with minimal fluctuations.

 






\begin{table}[!htbp]
	\caption{Contradistinction researchOur between our method and relatively good traditional machine learning algorithms on multiple indicators and stability analysis in the five OTU datasets.} 
	\resizebox{\linewidth}{!}{
	
    \fontsize{26}{30}\selectfont 

		\begin{tabular}{c|c|ccccccccc}
		\toprule
		\multirow{1}{*}{Metrics}  &\multirow{1}{*}{Method}  &\multirow{1}{*}{Cirrhosis}  &\multirow{1}{*}{IBD}  &\multirow{1}{*}{Obesity} &\multirow{1}{*}{T2D} &\multirow{1}{*}{WT2D}\\

		\midrule	
		\multirow{3}{*}{Accuracy $\uparrow$}    &RF		  &$0.877 \pm 0.043$  &$0.809 \pm 0.050$  &$0.644 \pm 0.028$     &$0.664 \pm 0.052$  &$0.703 \pm 0.105$    \\
									&SVM		  &$0.834 \pm 0.052$  &$0.809 \pm 0.066$  &$0.636 \pm 0.042$     &$0.613 \pm 0.057$  &$0.596 \pm 0.102$    \\
									&UMMAN(Ours)		  &$\textbf{0.886} \pm \textbf{0.006}$  &$\textbf{0.889} \pm \textbf{0.036}$  &$\textbf{0.706} \pm \textbf{0.033}$     &$\textbf{0.748} \pm \textbf{0.026}$  &$\textbf{0.867} \pm \textbf{0.001}$    \\
		
		\midrule
		\multirow{3}{*}{Precision $\uparrow$}   
								&RF		   &$0.890 \pm 0.041$  &$0.720 \pm 0.106$  &$0.540 \pm 0.109$     &$0.670 \pm 0.054$  &$0.730 \pm 0.114$    \\ 
								&SVM		   &$0.840 \pm 0.052$  &$0.780 \pm 0.097$  &$0.560 \pm 0.103$     &$0.620 \pm 0.060$  &$0.590 \pm 0.132$    \\
								&UMMAN(Ours)		   &$\textbf{0.894} \pm \textbf{0.056}$  &$\textbf{0.931} \pm \textbf{0.024}$  &$\textbf{0.721} \pm \textbf{0.101}$     &$\textbf{0.846} \pm \textbf{0.050}$  &$\textbf{0.914} \pm \textbf{0.070}$    \\

		\midrule
		\multirow{3}{*}{Recall $\uparrow$}    
							   &Recall		   &$0.880 \pm 0.044$  &$0.810 \pm 0.050$  &$0.640 \pm 0.028$     &$0.660 \pm 0.052$  &$0.700 \pm 0.105$    \\
							   &SVM		   &$0.830 \pm 0.052$  &$0.810 \pm 0.066$  &$0.640 \pm 0.042$     &$0.610 \pm 0.057$  &$0.600 \pm 0.102$    \\
							   &UMMAN(Ours)		   &$\textbf{0.952} \pm \textbf{0.001}$  &$\textbf{0.969} \pm \textbf{0.023}$  &$\textbf{0.758} \pm \textbf{0.048}$     &$\textbf{0.973} \pm \textbf{0.022}$  &$\textbf{0.857} \pm \textbf{0.001}$    \\
		
		\midrule
		\multirow{3}{*}{F1-Score $\uparrow$}    
								&RF		   &$0.880 \pm 0.045$  &$0.750 \pm 0.073$  &$0.540 \pm 0.038$     &$0.660 \pm 0.053$  &$0.690 \pm 0.109$    \\
							  	&SVM		   &$0.830 \pm 0.053$  &$0.780 \pm 0.076$  &$0.550 \pm 0.048$     &$0.6107 \pm 0.058$  &$0.570 \pm 0.112$    \\
								&UMMAN(Ours)		  &$\textbf{0.888} \pm \textbf{0.006}$  &$\textbf{0.925} \pm \textbf{0.023}$  &$\textbf{0.739} \pm \textbf{0.065}$     &$\textbf{0.732} \pm \textbf{0.037}$  &$\textbf{0.857} \pm \textbf{0.001}$    \\

		\midrule
		\multirow{3}{*}{AUC $\uparrow$}    
								&RF		   &$0.945 \pm 0.036$  &$0.890 \pm 0.078$  &$0.655 \pm 0.079$     &$0.744 \pm 0.056$  &$0.762 \pm 0.111$    \\
								&SVM		   &$0.922 \pm 0.041$  &$0.862 \pm 0.083$  &$0.648 \pm 0.071$     &$0.663 \pm 0.066$  &$0.664 \pm 0.126$    \\
								&UMMAN(Ours)		   &$\textbf{0.954} \pm \textbf{0.002}$  &$\textbf{0.930} \pm \textbf{0.018}$  &$\textbf{0.684} \pm \textbf{0.040}$     &$\textbf{0.777} \pm \textbf{0.027}$  &$\textbf{0.797} \pm \textbf{0.027}$    \\

		\bottomrule
		\end{tabular}
	}
            \label{table2}
	\end{table}

\subsection{ABLATION STUDY}
In our method, Attention block, Node Feature Global Integration (NFGI) descriptor and Adversarial loss are three core components, which is to improve the performance on the OTU datasets. We conduct an ablation study on eight variants: a) Ours (Base), only with the framework of the UMMAN model; b) Ours (Adv), with Adversarial loss which is used to extract the features of the Original-Graph; c) Ours (Attn), with Attention block which is used to get the embedding of the Original-Graph and the Shuffled-Graph. In this study, we use the global average pooling of Multi-Graphs to replace Attention block, the global average pooling is defined as:
\begin{equation}
  \mathbf{v} = GAP(\mathbf{\phi_{\theta}(X)}) \\ 
\end{equation}
d) Ours (NFGI), adopting Node Feature Global Integration(NFGI) which is designed to describe the global graph embedding, replace this module with the average value of each node in this part; e) Ours (Attn+Adv), with Attention block and Adversarial loss; f) Ours (NFGI+Adv), with NFGI and Adversarial loss; g) Ours (Attn+NFGI), with Attention block and NFGI. h) Ours (Full), with Attention block,  NFGI and Adversarial loss. The numeric comparisons on \textbf{\textit{Cirrhosis}}, \textbf{\textit{IBD}}, \textbf{\textit{Obesity}}, \textbf{\textit{T2D}}, \textbf{\textit{WT2D}} are shown in \hyperref[table3]{\textcolor{blue}{Table 3}}. On the whole, the method with Attention block,  NFGI and Adversarial loss, i.e., Ours (Full) performs the best. It is worth mentioning that even if our method is ablated, the effect on most indicators is better than previous work.

\begin{table}[!htbp] 
	\caption{Comparisons on the performance gains with the Attention Block and NFGI module in terms of two metrics.} 
	\resizebox{\linewidth}{!}{
   \fontsize{16}{22}\selectfont 

	\begin{tabular}{cccccccccccccc}
		\toprule
		\multirow{2}*{Method} & \multirow{2}*{ATTN} & \multirow{2}*{NFGI} & \multirow{2}*{ADV} & \multicolumn{2}{c}{Cirrhosis} & \multicolumn{2}{c}{IBD} & \multicolumn{2}{c}{Obesity} & \multicolumn{2}{c}{T2D} & \multicolumn{2}{c}{WT2D}\\
		\cmidrule(lr){5-6}\cmidrule(lr){7-8}\cmidrule(lr){9-10}\cmidrule(lr){11-12}\cmidrule(lr){13-14}
		& & & & Acc & AUC & Acc & AUC & Acc & AUC & Acc & AUC & Acc & AUC\\
		\midrule
		Ours (Base)  & \textcolor{red}{\ding{55}} & \textcolor{red}{\ding{55}} & \textcolor{red}{\ding{55}} & 0.832 & 0.918 & 0.809 & 0.781 & 0.644 & 0.606 & 0.670 & 0.727 & 0.697 & 0.725\\
		Ours (Adv) & \textcolor{red}{\ding{55}} & \textcolor{red}{\ding{55}} & \textcolor{red}{\Checkmark} & 0.885 & 0.945 & 0.832 & 0.823 & 0.676 & 0.636 & 0.695 & 0.727 & 0.721 & 0.705 \\
		Ours (Attn) & \textcolor{red}{\Checkmark} & \textcolor{red}{\ding{55}} & \textcolor{red}{\ding{55}} & 0.859 & 0.935 & 0.816 & 0.810 & 0.645 & 0.621 & 0.688 & 0.714 & 0.703 & 0.749 \\
		Ours (NFGI) & \textcolor{red}{\ding{55}} & \textcolor{red}{\Checkmark} & \textcolor{red}{\ding{55}} & 0.841 & 0.930 & 0.830 & 0.817 & 0.658 & 0.613 & 0.676 & 0.720 & 0.711 & 0.732 \\
		Ours (Attn+Adv) & \textcolor{red}{\Checkmark} & \textcolor{red}{\ding{55}} & \textcolor{red}{\Checkmark} & 0.882 & 0.933 & 0.818 & 0.844 & 0.657 & 0.676 & 0.685 & 0.715 & 0.733 & 0.773 \\
		Ours (NFGI+Adv)  & \textcolor{red}{\ding{55}} & \textcolor{red}{\Checkmark} & \textcolor{red}{\Checkmark} & 0.861 & 0.929 & 0.808 & 0.774 & 0.672 & 0.637 & 0.721 & 0.770 & 0.800 & 0.767 \\
		Ours (Attn+NFGI)  & \textcolor{red}{\Checkmark} & \textcolor{red}{\Checkmark} & \textcolor{red}{\ding{55}} & 0.867 & 0.934 & 0.836 & 0.826 & 0.651 & 0.626 & 0.724 & 0.763 & 0.764 & 0.768 \\
		Ours (Full)  & \textcolor{red}{\Checkmark} & \textcolor{red}{\Checkmark} & \textcolor{red}{\Checkmark} & \textbf{0.886} & \textbf{0.954} & \textbf{0.889} & \textbf{0.930} & \textbf{0.706} & \textbf{0.684} & \textbf{0.748} & \textbf{0.777} & \textbf{0.867} & \textbf{0.797} \\
		\bottomrule
	\end{tabular}
	}
 \label{table3}
\end{table}

\begin{table}[!htbp] 
	\caption{Comparisons on the performance gains with the Attention Block and NFGI module in terms of two metrics.} 
	\resizebox{\linewidth}{!}{
   \fontsize{16}{22}\selectfont 

	\begin{tabular}{ccccccccccccccc}
		\toprule
		\multirow{1}*{Method} & \multirow{1}*{DAN} & \multirow{1}*{MCC} & \multirow{1}*{KD} & \multirow{1}*{S1} & \multirow{1}*{S2} & \multirow{1}*{S3} & \multirow{1}*{S4} & \multirow{1}*{S5} & \multirow{1}*{S6} & \multirow{1}*{S7} & \multirow{1}*{S8} & \multirow{1}*{S9} & \multirow{1}*{Avg.}\\
		\midrule
		Ours (Base)  & \textcolor{red}{\ding{55}} & \textcolor{red}{\ding{55}} & \textcolor{red}{\ding{55}} & 66.528 & 55.618 & \textbf{57.666} & 84.919 & 74.540 & 68.695 & 67.945 & 75.842 & 70.583 & $69.148\pm{0.696}$\\
		Ours (DAN) & \textcolor{red}{\Checkmark} & \textcolor{red}{\ding{55}} & \textcolor{red}{\ding{55}} & 65.667 & 55.853 & 57.167 & 86.270 & 74.730 & 69.972 & 70.444 & 75.921 & 70.556 & $69.620\pm{0.506}$\\
		Ours (MCC)  & \textcolor{red}{\ding{55}} & \textcolor{red}{\Checkmark} & \textcolor{red}{\ding{55}} & 63.444 & 55.176 & 54.472 & 91.946 & 77.946 & 74.333 & 73.472 & 76.158 & 67.917 & $70.541\pm{0.570}$\\
		Ours (DAN+MCC)  & \textcolor{red}{\Checkmark} & \textcolor{red}{\Checkmark} & \textcolor{red}{\ding{55}} & 63.000 & 55.765 & 56.083 & 92.378 & 77.378 & 73.694 & 73.778 & 76.079 & 72.278 & $71.159\pm{0.387}$\\
		Ours (Full)  & \textcolor{red}{\Checkmark} & \textcolor{red}{\Checkmark} & \textcolor{red}{\Checkmark} & \textbf{69.944} & \textbf{57.794} & 57.056 & \textbf{93.946} & \textbf{86.270} & \textbf{79.583} & \textbf{76.472} & \textbf{76.842} & \textbf{77.944} & $\textbf{75.095} \pm \textbf{0.305}$\\
		\bottomrule
	\end{tabular}
        }
 \label{table4}
\end{table}

\section{CONCLUSION}
In this paper, we propose a novel architecture UMMAN which combines GNN with disease prediction tasks based on intestinal flora for the first time. It helps learn the multiplex connection among gut microbes of different hosts. We construct the Multi-Graph and Shuffled-Graph with multiple relation-types, get the embedding of hosts through improved GCN. In addition, we introduce the Node Feature Global Integration (NFGI) to describe the global embedding of a graph with node-level stage and graph-level stage. Finally, we design a joint loss consisting of adversarial loss and hybrid attention loss as the final loss function. Extensive experiments show that our UMMAN performs well on the task of disease prediction based on intestinal flora and achieves the state-of-the-art. Our UMMAN model can be applied to assist in the diagnosis of gut microbiome-related diseases.

In the future, the main work may be to find a more suitable way to construct a gut microbiota, in order to avoid building microbial interaction networks incorrectly, which could negatively impact the accuracy of subsequent graph classification tasks. Lastly, the biggest challenge in this study is the data, as there are currently relatively few publicly available gut microbiome datasets for disease classification. However, with the development of sequencing technology, more datasets that can reflect sample features may emerge in the future. At that time, the model's performance may improve after extensive training.

\vfill

\end{document}